%% file: main.tex
\pdfoutput=1
\documentclass[11pt]{article}

\usepackage[]{EMNLP2023}

\usepackage{times}
\usepackage{latexsym}

\usepackage[T1]{fontenc}
\usepackage[utf8]{inputenc}

\usepackage{microtype}

\usepackage{inconsolata}
\usepackage{bbding}
\usepackage{graphicx}
\usepackage{subfigure}
\usepackage{wrapfig}
\usepackage{makecell}
\usepackage{multirow}
\usepackage{lineno,hyperref}
\usepackage{hyperref}
\usepackage{url}
\usepackage{geometry}
\usepackage{color}
\usepackage{float}
\usepackage{longtable}
\usepackage{graphicx}
\usepackage{booktabs}
\usepackage{times}
\usepackage{latexsym}
\usepackage{makecell}
\usepackage{array, tabularx, caption, boldline}
\usepackage{graphicx}
\usepackage{cellspace}
\usepackage{multirow}
\usepackage{color}
\title{Harnessing the Power of David against Goliath: Exploring Instruction Data Generation without Using Closed-Source Models}

\author{
    Yue Wang\textsuperscript{\normalfont 1,2},
    Xinrui Wang\textsuperscript{\normalfont 1},
    Juntao Li\textsuperscript{\normalfont 1}\thanks{\; Corresponding author. Work is done during the internship of Yue Wang at Ant Group.},
    Jinxiong Chang\textsuperscript{\normalfont 2},\\
    {\bf Qishen Zhang}\textsuperscript{\normalfont 2},
    {\bf Zhongyi Liu}\textsuperscript{\normalfont 2},
    {\bf Guannan Zhang}\textsuperscript{\normalfont 2},
    {\bf Min Zhang}\textsuperscript{\normalfont 1}
    \\ 
    \textsuperscript{1}School of Computer Science and Technology, Soochow University\\
    \textsuperscript{2}Ant Group\\
  \texttt{ywangnlp@stu.suda.edu.cn} \\
}

\begin{document}
\maketitle
\begin{abstract}
%which necessitates diverse instructions, factual responses, and alignment between instructions and responses.
% We first investigate different variants of existing instruction generation methods and then combine the most effective variant with our two novel strategies to further improve the quality.
Instruction tuning is instrumental in enabling Large Language Models~(LLMs) to follow user instructions to complete various open-domain tasks.
The success of instruction tuning depends on the availability of high-quality instruction data.
Owing to the exorbitant cost and substandard quality of human annotation, recent works have been deeply engaged in the exploration of the utilization of powerful closed-source models to generate instruction data automatically.
However, these methods carry potential risks arising from the usage requirements of powerful closed-source models, which strictly forbid the utilization of their outputs to develop machine learning models.
To deal with this problem, in this work, we explore alternative approaches to generate high-quality instruction data that do not rely on closed-source models.
Our exploration includes an investigation of various existing instruction generation methods, culminating in the integration of the most efficient variant with two novel strategies to enhance the quality further. 
Evaluation results from two benchmarks and the GPT-4 model demonstrate the effectiveness of our generated instruction data, which can outperform Alpaca, a method reliant on closed-source models.
We hope that more progress can be achieved in generating high-quality instruction data without using closed-source models.

\end{abstract}

\input{article/introduction}
\input{article/study}
\input{article/method}
\input{article/experiment}
\input{article/relatedwork}
\input{article/conclusion}

\section*{Limitations}
Although our instruction framework can generate high-quality instruction data, it still has the following limitations:
\begin{itemize}
    \item \textbf{Hallucinations}: Even though our generated data utilize the existing corpus as outputs to ensure the validity of outputs, due to the characteristic of the used backbone LLMs, it may still have the hallucination problem;
    \item \textbf{Instruction Diversity}: Despite our proposed strategy, the diversity of generated instruction can be further improved.
    Through the diversity analysis, the generated instruction data is still far from satisfactory.

\end{itemize}

\section*{Ethics Statement}

The instruction data are generated automatically from LLMs, which do not represent the viewpoints of the authors.
Due to social bias and lack of professional knowledge, the generated data may contain misleading and toxic information, which needs to be addressed before being applied to realistic scenarios.
To promote the development of instruction generation without using closed-source models, we will release our codes and generated data.

% \section*{Acknowledgements}

\bibliography{main}
\bibliographystyle{acl_natbib}

\clearpage

\appendix

\section{The Diversity of SUPERNI and LongForm}
To compare the diversity of instructions, we also visualize the statistics of SUPERNI and LongForm test sets, which is shown in Figure~\ref{fig:diverse_test}.
\input{figure/fig_benchmark}

\section{The Templates of Different Strayegies}
We show the instruction following prompt of Alpaca in Table~\ref{tab:alpaca_insturction}, the instruction data generation prompt of Alpaca in Table~\ref{tab:alpacainstruction}, the instruction data generation prompt of LongForm in Table~\ref{tab:longform_generation} and the LLM extraction prompt in Table~\ref{tab:llmextract}.

\input{table/prompt}

\input{table/prompt1}

\input{table/prompt2}

\input{table/table_extractprompt}

\end{document}

%% file: article/introduction.tex
\section{Introduction}
% aims to make Large Language Models~(LLMs) generate outputs aligned with the intent of user instructions, which
Instruction tuning has received a wide range of attention from the Natural Language Processing~(NLP) community~\citep{mishra2021cross,wei2021finetuned,sanh2021multitask,chung2022scaling,ouyang2022training,wang2022self}.
 Through the utilization of various types of instruction data, LLMs can fully exploit the knowledge obtained in the pre-training stage, leading to the superior performance of generalization capability across unseen tasks. 
 This ability to transcend task-specific limitations illustrates the potential of LLMs for Artificial General Intelligence (AGI).
 
\input{figure/fig_pipeline}

The quality of data is paramount to the success of instruction tuning~\citep{zhou2023lima,chen2023maybe}.
Existing instruction datasets fall into two categories: those created through manual annotation and those generated by models.
The former method involves the manual conversion of existing NLP datasets into an instruction format, but this approach is constrained by high cost and variable quality~\citep{wang2022self,honovich2022unnatural}.
The latter method existing methods, on the other hand, typically use powerful closed-source models to generate instruction datasets~\citep{wang2022self,honovich2022unnatural,alpaca,koksal2023longform,yin2023dynosaur}.
Unfortunately, the usage requirements of powerful closed-source models usually restrict using their outputs for developing machine learning models, e.g., OpenAI~\footnote{\url{https://openai.com/policies/terms-of-use}}, Google~\footnote{\url{https://policies.google.com/terms/generative-ai}} and Anthropic~\footnote{\url{https://vault.pactsafe.io/s/9f502c93-cb5c-4571-b205-1e479da61794/legal.html\#terms}}, introducing potential risks to existing model generation methods.
Therefore, the development of a novel method to automatically generate high-quality instruction data without relying on closed-source models becomes imperative. 
However, fulfilling this demand is challenging due to the need for high-quality instruction data that meets multiple criteria including instruction diversity, output validity, and tight alignment between instruction and output.
Due to the limited capabilities of open-source models, it is more challenging to address these problems.

To address this problem, in this work, we first explore different variants of the existing instruction data generation methods when do not use closed-source models.
We then introduce a novel framework for instruction data generation that does not rely on closed-source models, as shown in Figure~\ref{fig:model}.
Our framework consists of three main components: the training of the instruction generation model, the generation of the instruction data, and the filtering of generated instructions. 
Specifically, we first explore the effectiveness of different variants of existing instruction generation methods without using closed-source models.
Subsequently, to enhance the alignment between generated instructions and selected outputs, we propose a novel instruction filtering strategy to select the most appropriate instruction from a pool of generated candidate instructions.
Moreover, to improve the diversity of instruction, we introduce a novel extract-then-generate strategy, which extracts varied segments from the existing passages to produce more diverse instructions.
Experimental results on two instruction tuning benchmarks confirm the effectiveness of our generated data, which outperform total nine instruction datasets, including in-domain datasets and those generated by closed-source models. 
Further evaluations based on GPT-4 demonstrate our data can slightly exceed the performance of Alpaca.

In conclusion, the contributions of this work are as follows:
\begin{itemize}
    \item We first explore the potential of generating instruction data without using closed-source models by studying different variants of existing instruction data generation methods;
    \item We propose a novel instruction filtering strategy to ensure the alignment between the generated instructions and outputs.
    To improve the diversity of instructions, we also propose an extract-then-generate strategy;
    \item Benchmark and GPT-4 evaluation results confirm the effectiveness of open-source models in generating high-quality instruction data, delivering results on par with closed-source models generated datasets.
\end{itemize}
% in utilizing the data sources used for training instruction generation models and generating instruction
% introduces a novel instruction generation framework that ensures instruction diversity, output factuality, and strong relevance between instruction and output without relying on closed-source models. 
% The core idea of our framework is to expand existing instruction data, which is non-competitive with the automatically generated data using OpenAI's API. 

%% file: figure/fig_pipeline.tex
\begin{figure}[t]
\begin{center}
\includegraphics[width=\columnwidth]{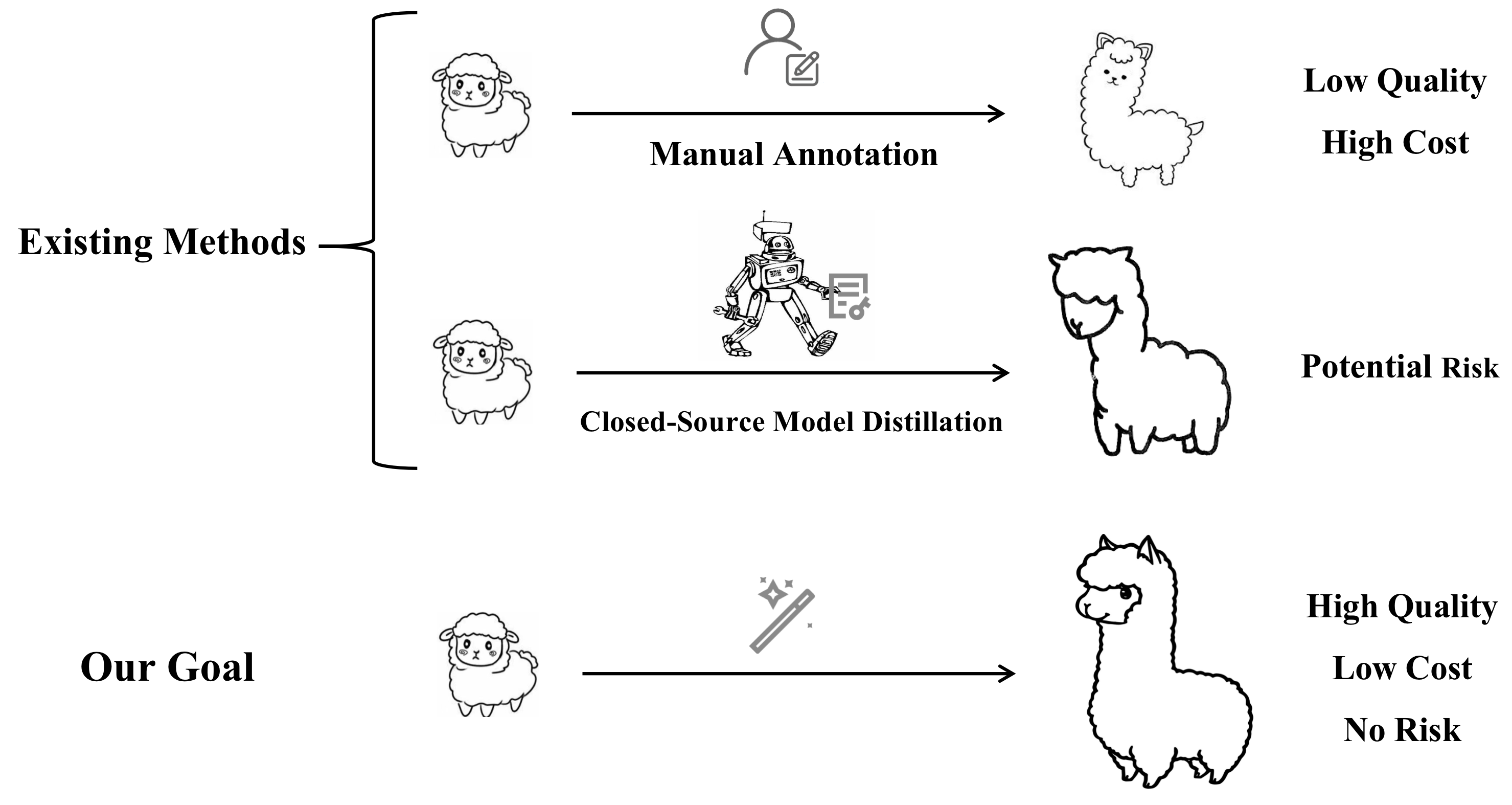}

\end{center}
\caption{\label{fig:pipiline} Comparison of different instruction data generation methods.
We focus on generating high-quality instruction cheaply without using closed-source models.
}
\end{figure}

%% file: article/study.tex
\section{Study on Adapting Existing Instruction Generation Approaches}
\input{table/table_length}
\input{figure/fig_model}

In this section, we first introduce the experimental settings and then explore some variants of existing instruction generation methods when do not use closed-source models.
% Specifically, we study the different data sources and the effect of data numbers.
% explore the potential of existing instruction generation methods when do not use closed-source models.
% Specifically, in addition to adapting existing methods to generate instruction data by using open-source models directly, we also explore some variants, including train data source and inference data source.

\subsection{Implementation  Details}
We implement all models with the open-source toolkit \textit{Transformers}\footnote{\url{https://github.com/huggingface/transformers}}~\cite{wolf2020transformers}.
In the fine-tuning stage, we follow \citet{alpaca} use AdamW~\citep{loshchilov2018decoupled} optimizer and set the learning rate to 2e-5, batch size to 128, learning rate warmup ratio to 0.03, and perform training for 3 epochs.
We utilize LLaMA-7B~\citep{touvron2023llama} as the backbone model and update all parameters during the fine-tuning stage. 
All the fine-tuning experiments are performed with 80GB NVIDIA A100 GPUs.
During the instruction generation stage, we employ Nucleus Sampling~\citep{holtzman2019curious} to generate diverse instruction data.
Specifically, we set the Top-p to 0.9, the Top-k to 40, and the temperature to 0.7.

\subsection{Automatic Evaluation Benchmarks}
To evaluate different instruction methods automatically and comprehensively, we select two popular benchmarks with different characteristics: SUPER-NATURALINSTRUCTIONS~(SUPERNI)~\citep{wang2022super} and LongForm~\citep{koksal2023longform}.
We show the statistic of two benchmarks in Table~\ref{tab:data}.
SUPERNI is a large-scale instruction tuning benchmark, encompassing 1,616 NLP tasks.
To ensure consistency, we use the official test set splitting~\citep{wang2022super}. 
The test set consists of 15,310 instances, categorized into 12 task categories and 154 tasks in total. 
These tasks include a wide range of classification and generation tasks at the word, sentence, and document levels, with a maximum of 100 instances per task.
LongForm is a long text generation instruction benchmark, which consists of 2,045 examples and is generated from OpenAI `text-davinci-003' model according to English Wikipedia or C4 corpus, or natural instruction corpus~(Stack Exchange~\citep{gao2020pile} and WikiHow~\citep{koupaee2018wikihow}) or existing NLP tasks~(BigBench~\citep{srivastava2022beyond}, Enron~\citep{enron}, BEA-2019 shared task on grammatical error correction~\citep{bryant2019bea}).
For the automatic metrics,  we use the original evaluation metrics~\citep{wang2022super,koksal2023longform}, i.e., ROUGE-L~\citep{lin2004rouge} score for SUPERNI and METEOR~\citep{banerjee2005meteor} score for LongForm.
We show the statistic of two benchmarks in Table~\ref{tab:data}.
We can find that due to the different task types, the two benchmarks differ significantly in the average lengths.

\subsection{Instruction Dataset Baselines}

To conduct a comprehensive comparison, we compare various instruction datasets: 
\textbf{(1)~Self-Instruct}~\citep{wang2022self} employs vanilla GPT3~\citep{brown2020language} to generate instruction data automatically;
\textbf{(2)~Alpaca}~\citep{alpaca} introduces refinements to the pipeline of Self-Instruct to generate more high-quality data;
\textbf{(3)~LLaMA-GPT4}~\citep{peng2023instruction} deftly leverages the formidable power of GPT-4~\citep{openai2023gpt4} in order to generate responses for the instructions of Alpaca adeptly;
\textbf{(4)~Evol-Instruct}~\citep{xu2023wizardlm} put forth some prompt engineering strategies aimed at enhancing the complexity of instructions;
\textbf{(5)~Dromedary}~\citep{sun2023principle} propose elaborate prompt engineer strategies to generate instruction data with the use of LLaMa-65B~\citep{touvron2023llama};
\textbf{(6)~DYNASAUR}~\citep{yin2023dynosaur}  leverages the power of LLMs to generate instruction data by harnessing the metadata and data fields of existing NLP datasets;
\textbf{(7)~Dolly}~\citep{dolly} stands as a remarkable open-source dataset, meticulously composed of human-written instructions that cater to a wide range of general-purpose tasks.
We also compare with the training split of \textbf{(8)~SUPERNI}~\citep{wang2022super} and \textbf{(9)~LongForm}~\citep{koksal2023longform}) to compare with in-domain datasets.

\input{table/table_main}

\subsection{Results}
In Table~\ref{tab:main}, we find that almost all existing instruction datasets cannot achieve good performance on both benchmarks.
We think this phenomenon results from the significantly different statistics of the two benchmarks, which need models to master different abilities.
Besides, both SUPERNI and LongForm only bring significant improvements on the corresponding test set, while bringing little improvements on the other test set.
This phenomenon further confirms our hypothesis that it requires different skills to deal with SUPERNI and LongForm.
% Additionally, we f
% Besides, we can find that, with the use of close-sourced models, the instruction dataset can achieve more balance results on two benchmarks, which shows the power of close-sourced models.
% Due to the unsatisfactory performance of all instruction g

\subsection{Exploring the Potential of Existing Instruction Generation Methods}
Due to the usage requirements, it is forbidden to use closed-source models to generate instruction data.
Therefore, we explore the potential of existing instruction generation methods when do not use closed-sourced models.

\paragraph{Instrution Data Generation Methods}
We first try to adapt existing instruction data generation methods. 
Specifically, we study two typical methods: \textbf{Alpaca}~\citep{touvron2023llama} and \textbf{LongForm}~\citep{koksal2023longform}. 
Alpaca generates instruction data in a step-by-step manner, starting with the generation of instructions, followed by the generation of inputs based on these instructions, and ultimately producing outputs by seamlessly combining the instructions with the corresponding inputs. 
LongForm, on the other hand, first collects passages from the existing corpus and uses closed-source LLMs to generate the instruction for these selected passages. 
In our preliminary experiments, we observe that both Alpaca and LongForm encounter challenges in generating high-quality instruction data with in-context learning when using LLaMa-7B as backbone models. 
We attribute this difficulty to the discernible ability gap between LLaMa-7B and ChatGPT.
To address this problem, we leverage existing instruction data to train a specialized instruction data generator.
Specifically, we transform the existing instruction data into the instruction generation templates of Alpaca and LongForm to use training data to update the parameters of LLaMa-7B.
The templates used are shown in the Appendix.
We ensure that the data used to train the instruction data generator is not collected from the outputs of closed-source models. 
% This allows us to stimulate the instruction generation ability by fine-tuning the parameters of LLaMa-7B, resulting in a specialized instruction data generator. 
However, despite updating the parameters to develop a specific instruction data generator, Alpaca still has difficulty generating high-quality data.
After rule-based data processing, the generation speed of Alpaca is limited to a mere 30 instances per hour on 1x 80GB NVIDIA A100 GPU. 
Therefore, in the following experiments, we follow the approach of LongForm to generate instruction data.

\paragraph{Training Strategy}
We explore the effect of different training strategies of the instruction generator, including instruction formats, training data source and size.
Specifically, we first study whether the open-source instruction following model can generate high-quality instruction data.
We show the details of instruction generation and the following formats in the Appendix.
We also select three training data sources: 
\textbf{(1)~Seed Instructions}: is introduce by Self-Instruct~\citep{wang2022self}, which contains 175 human written instruction data;
\textbf{(2)~Dolly}~\citep{dolly} is a human-written general-purpose instruction dataset;
\textbf{(3)~SUPERNI}~\citep{wang2022super}.

\paragraph{Inference Strategy}
We also explore the effect of the inference data sources.
Specifically, we collect fragments from both unsupervised and supervised corpus.
For unsupervised corpus, we follow \citet{koksal2023longform} to sample 9,000 passages from the Wikipedia corpus and 4,500 passages from the C4 corpus.
For supervised corpus, we sample data from SUPERNI~\citep{wang2022super}.
\input{table/table_train_source_unsupervised}

\input{table/table_train_source_supervised}

\paragraph{Results and Analysis}
We report the results of different training and inference strategies in Table~\ref{tab:train_source_unsupervised} and ~\ref{tab:train_source_supervised}.
From the results in Table~\ref{tab:train_source_unsupervised}, we can find that with only 175 instruction data, LLaMA-7B can generate better instructions than OpenAI `text-davinci-003’ model for unsupervised corpus.
Table~\ref{tab:train_source_supervised} shows that the instruction data generator has difficulty in generating high-quality instruction data for supervised corpus.
We argue that the statistics of different corpus cause this phenomenon.
The outputs of the supervised corpus are usually very short, especially for answer choice tasks, where the answer is only one character, `A' or 'B', which only provides limited information to the model and is difficult to generate diverse high-quality data.
Therefore, only with the help of a filtering strategy, combined with 13,500 training data, can achieve comparable performance to manual annotation.
In the next section, we will introduce the instruction filtering strategy in detail.

%% file: table/table_length.tex
\begin{table}[!tbp]
\centering
\renewcommand\arraystretch{1.1}
\resizebox{\columnwidth}{!}{
    \begin{tabular}{l  c c c c c c}
        \hline
        \bf Dataset &\bf Number& \bf Avg. Ins. Len.& \bf Avg. Out. Len. &\bf Task Type \\
        \hline
        SUPERNI& 11,810&186.62&5.86&Classification\&Generation\\
        LongForm&2,045 &129.94&300.86&Long Text Generation\\
        \hline

    \end{tabular}
}
\caption{\label{tab:data}The statistics of the SUPERNI and LongForm test sets. 
\textbf{Avg. Ins. Len.} denotes the average length of instructions.
\textbf{Avg. Out. Len.} denotes the average length of golden outputs.
}
% \vspace{-0.4cm}
\end{table}

%% file: figure/fig_model.tex
\begin{figure*}[htbp]
\begin{center}
\includegraphics[width=0.9\textwidth]{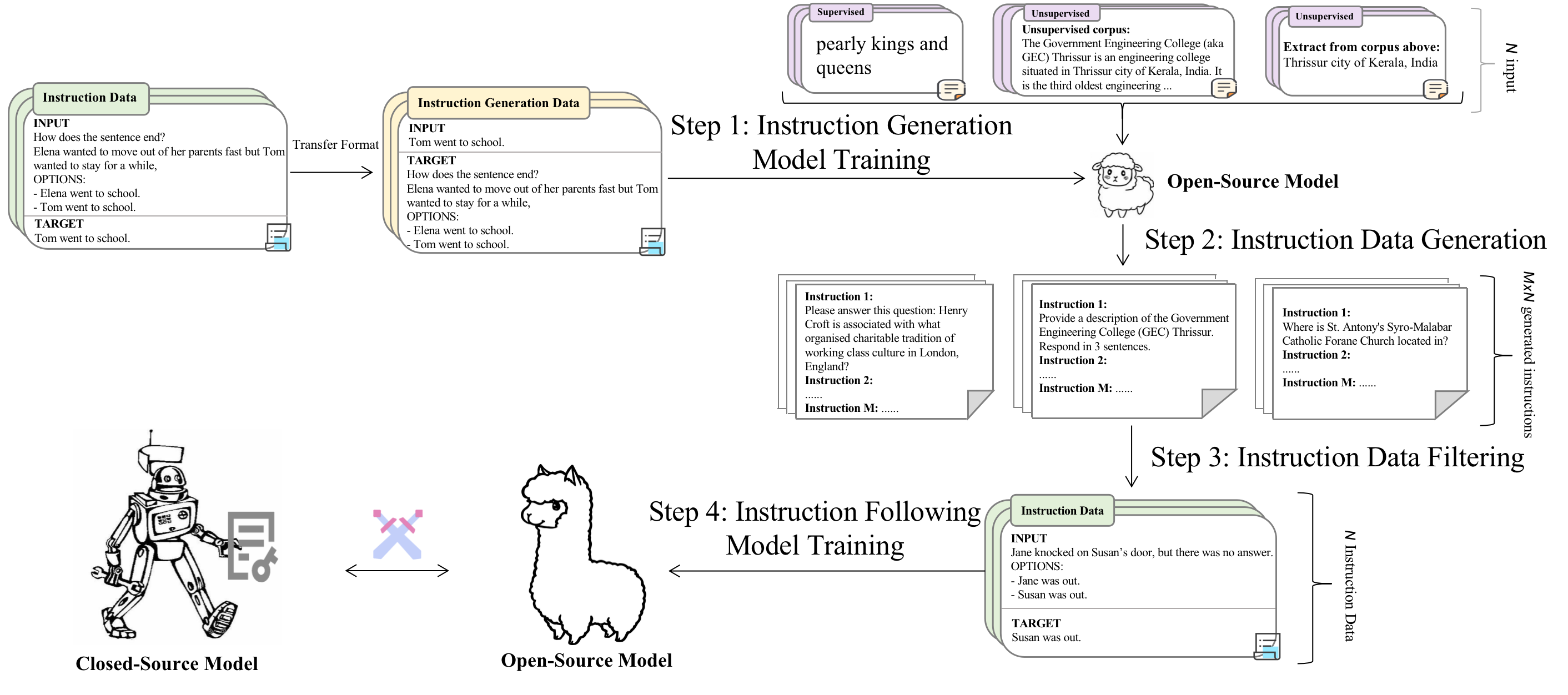}

\end{center}
\caption{\label{fig:model} An illustration of our instruction data generation framework, which does not use closed-source models throughout the process.
Specifically, in \textbf{Step 1}, we reverse the instruction and output of original instruction data to train an instruction generation model;
in \textbf{Step 2}, we use the fragments collected from the existing corpus as the output and use the instruction generation model to generate candidate instructions for these selected outputs;
in \textbf{Step 3}, we propose a novel instruction filtering strategy to select the most appropriate instruction;
in \textbf{Step 4}, we use the filtered instruction data to train the final instruction following model.
}

\end{figure*}

%% file: table/table_main.tex
\begin{table*}[!tbp]
\centering

\resizebox{0.9\textwidth}{!}{
\begin{tabular}{l c c c c c }
\\
%\toprule[1.25pt]
\toprule
\bf Method	&\bf Potential Risks&\bf Data Size& \bf SUPERNI & \bf LongForm\\
 \midrule
LLaMA-7B~\citep{touvron2023llama}&\XSolidBrush&-&9.88&13.2\\
\quad +SELF-INSTRUCT~\citep{wang2022self}&\Checkmark&82K&37.58&10.5\\
\quad +Alpaca~\citep{alpaca}&\Checkmark&52K&38.62&14.06\\
\quad +LLaMA-GPT4~\citep{peng2023instruction}&\Checkmark&52K&30.22&15.78\\
\quad +Evol-Instruct~\citep{xu2023wizardlm}&\Checkmark&70K&21.36&16.9\\
\quad +DYNASAUR~\citep{yin2023dynosaur}&\Checkmark&52K&41.06&12.60\\
\quad +LongForm~\citep{koksal2023longform}&\Checkmark&24K&18.18&18.55\\
\quad +Dolly~\citep{dolly}&\XSolidBrush&15K&20.10&13.5\\
\quad +Dromedary~\citep{sun2023principle}&\XSolidBrush&52K&11.56&18.70\\
\quad +SUPERNI~\citep{wang2022super}&\XSolidBrush&75K&44.66&7.7\\
\quad +SUPERNI~\citep{wang2022super}+LongForm~\citep{koksal2023longform}&\Checkmark&28K&35.83&14.01\\

\hline
\quad +Ours~(Generated from Unsupervised Corpus)&\XSolidBrush&14K&8.82&\textbf{21.00}\\
\quad +Ours~(Generated from Supervised Corpus)&\XSolidBrush&14K&\textbf{45.00}&9.77\\
\quad +Ours~(Generated from Unsupervised +Supervised Corpus)&\XSolidBrush&28K&\underline{44.88}&\underline{18.84}\\
\bottomrule
\end{tabular}
}
\caption{\label{tab:main}  
The performance of our generated instruction data and all instruction dataset baselines on SUPERNI and LongForm benchmarks.
\textbf{\textit{Bold}} and \textit{\underline{Underline}} indicate the best and the second best performance respectively.
\textbf{Potential Risks} denotes whether use closed-source models to generate instructions, which have potential risks due to the terms of use.
Following the original  work~\citep{wang2022super,koksal2023longform}, we report the ROUGE-L~\citep{lin2004rouge} score for SUPERNI and METEOR~\citep{banerjee2005meteor} score for LongForm. 
}
% \vspace{-0.4cm}
\end{table*}

%% file: table/table_train_source_unsupervised.tex
\begin{table}[!tbp]
\centering

\resizebox{\columnwidth}{!}{
\begin{tabular}{c c c| c c }
\\
%\toprule[1.25pt]
\toprule
\multicolumn{3}{c|}{\textbf{Instruction Generation Model}} &  \multirow{2}{*}{\textbf{SUPERNI}} &  \multirow{2}{*}{\textbf{LongForm}}\\

 \textbf{Training Strategies}&\bf Training Data Source&\bf Training Data Size\\
 \hline
  \multicolumn{5}{c}{\textit{Automatic Generation with Using Open-Source Models}}\\
  \midrule
Instruction Generation & Seed Instruction&175&9.61&\textbf{18.01}\\
Instruction Generation & Dolly&14,563&10.00&\underline{17.01}\\
Instruction Generation & SUPERNI&13,500&5.95&16.24\\
Instruction Generation & Dolly+SUPERNI&28,063&\textbf{10.45}&16.59\\
Instruction Following & Dolly+SUPERNI&28,063&5.89&16.79\\
 % \hline
 %  \multicolumn{5}{c}{\textit{Automatic Generation With using Open-Source Models and Instrution Filtering Strategy}}\\
 %  \hline
 %  Instruction&175&8.82&\textbf{21.00}\\
 \hline
  \multicolumn{5}{c}{\textit{Automatic Generation with Using Closed-Source Models}}\\
  \hline
-&-&-&\underline{10.28}&16.15\\
\bottomrule
\end{tabular}
}
\caption{\label{tab:train_source_unsupervised}  
The performance of our generated instruction data on SUPERNI and LongForm benchmarks.
We generate instructions for 13,500 unsupervised passages collected from Wikipedia and C4 corpus.
\textbf{\textit{Bold}} and \textit{\underline{Underline}} indicate the best and the second best performance respectively.
For comparison, we also report the performance of using the OpenAI `text-davinci-003' model to generate instructions for the same passages.
}
% \vspace{-0.4cm}
\end{table}

%% file: table/table_train_source_supervised.tex
\begin{table}[!tbp]
\centering

\resizebox{\columnwidth}{!}{
\begin{tabular}{c c| c| c c c c }
\\
%\toprule[1.25pt]
\toprule
 \multicolumn{2}{c|}{\textbf{Instruction Generation Model}}  & \multirow{2}{*}{\textbf{Generated Data Size}}&  \multirow{2}{*}{\textbf{SUPERNI}}& \multirow{2}{*}{\textbf{LongForm}}\\
\bf Training Data Source&\bf  Training Data Size&\\
 \hline
  \multicolumn{5}{c}{\textit{Automatic Generation without Instruction Filtering Strategy}}\\
  \midrule
   % Seed ALL&350&13,500&17.25&5.73\\
    % All&29,300&13,500&39.96&8.29\\
 Seed Instruction&175&13,500&18.36&4.00\\
  Dolly&14,563&13,500&23.92&7.26\\
 SUPERNI&175&13,500&17.62&5.56\\
SUPERNI&13,500&13,500&40.34&\underline{9.50}\\
  SUPERNI&13,500&73,256&37.97&8.63\\
 \hline
  \multicolumn{5}{c}{\textit{Automatic Generation with Instruction Filtering Strategy}}\\
  \midrule
  SUPERNI&13,500&13,500&\textbf{45.00}&\textbf{9.77}\\
  SUPERNI&13,500&73,256&41.69&9.06\\
 \hline
  \multicolumn{5}{c}{\textit{Manual Annotation}}\\
  \hline
-&-&13,500&44.03&8.05\\
-&-&75,317  &\underline{44.66}&7.74\\
\bottomrule
\end{tabular}
}
\caption{\label{tab:train_source_supervised}  
The performance of our generated instruction data on SUPERNI and LongForm benchmarks
We generate instructions for supervised outputs collected from SUPERNI.
\textbf{\textit{Bold}} and \textit{\underline{Underline}} indicate the best and the second best performance respectively.
For comparison, we also report the performance of the original SUPERNI manual annotation.
}
% \vspace{-0.4cm}
\end{table}

%% file: article/method.tex
\section{Exploring Novel Strategies}

\input{table/table_filter}

\subsection{Instruction Filtering}
\paragraph{Strategy} In order to balance the diversity of instruction and the alignment between instruction and output, we propose a novel Instruction Filtering Strategy.
The motivation of this strategy is that the most appropriate instruction can infer the output as the max possible.
Specifically, we first generate multiple candidate instructions for one output.
Then we use the instruction following model to calculate the perplexity~(PPL) of the output given the candidate instruction.
Formally, we select the instruction that:
% \begin{equation}
${\arg\min}_I~ppl(O|I)$, where $I$ denotes the candidate instruction and $O$ denotes the passages selected as outputs.
% \end{equation}

\input{figure/fig_extract}

\paragraph{Results}
We first study the effect of different filtering models.
From the results in Table~\ref{tab:filter}, we can find that using the instruction filtering strategy has a significant effect on the performance, while the effect of different filtering models is not huge.
We think these results come from the fact that discrimination is simple than generative.
Therefore, to simplify the progress, in all the following experiments, we choose to use the original LLaMA-7B model as the filtering model.

\subsection{Extract-then-Generate}
\paragraph{Strategy} From the above results, we can find that the selected outputs have a large effect on the quality of the generated instruction.
Due to the single distribution of the outputs we use before that Wikipedia and C4 are too long and SUPERNI is too short, we try to diversify the distribution of outputs.
Because the output of SUPERNI is collected from existing NLP tasks, which means that they have been extracted manually, we try to extract parts from the Wikipedia and C4 passages.
We propose three strategies to extract passages:
\begin{itemize}
    \item \textbf{Keywords}: We use the toolkit Yake~\footnote{\url{https://github.com/LIAAD/yake}}~\citep{campos2018text,campos2018yake,campos2020yake} to extract the most unique keywords from the original passage;
    \item  \textbf{Random Sentence}: We extract one sentence from the original passage randomly;
    \item \textbf{LLM Extraction}: We try to utilize LLMs to extract the most valuable fragments from the original passage.
    Specifically, we use the LLaMA-7B-Dolly model to extract fragments. 
    We show the templates used for LLM Extraction in the Appendix.
\end{itemize}

\paragraph{Results}
We report the results of all the extraction strategies in Figure~\ref{fig:extract}.
We can find that compared to the original passages, the LLM extraction strategy brings significant improvements to SUPERNI and outperforms the other two methods.
There is a large amount of data in the unsupervised corpus, and how to efficiently use unsupervised data is still under-explored.
We hope the extract-then-generate strategy can shed light on generating diverse instruction data from unsupervised data.

%% file: table/table_filter.tex
\begin{table}[!tbp]
\centering

\resizebox{\columnwidth}{!}{
\begin{tabular}{c l c c }
\\
%\toprule[1.25pt]
\toprule
\bf Filtering&\bf Filtering Model	& \bf SUPERNI & \bf LongForm\\
 \midrule
\XSolidBrush&\quad \quad -&7.61&18.01\\
\hline
\Checkmark&LLaMa-7B&8.77&\underline{21.00}\\
\Checkmark&\quad +Dolly&8.34&20.85\\
\Checkmark&\quad +SUPERNI&8.50&20.91\\
\Checkmark&\quad +Seed Instrcution&\underline{8.82}&\textbf{21.12}\\
\Checkmark&\quad +All&\textbf{8.85}&20.97\\
\bottomrule
\end{tabular}
}
\caption{\label{tab:filter}  
The performance of our different filtering strategies on SUPERNI and LongForm benchmarks.
We use the instruction generation model trained with seed instructions to generate candidate instructions for 13,500 unsupervised passages collected from Wikipedia and C4 corpus.
\textbf{\textit{Bold}} and \textit{\underline{Underline}} indicate the best and the second best performance respectively.
}
% \vspace{-0.4cm}
\end{table}

%% file: figure/fig_extract.tex
\begin{figure}[t]
\begin{center}
\includegraphics[width=0.5\textwidth]{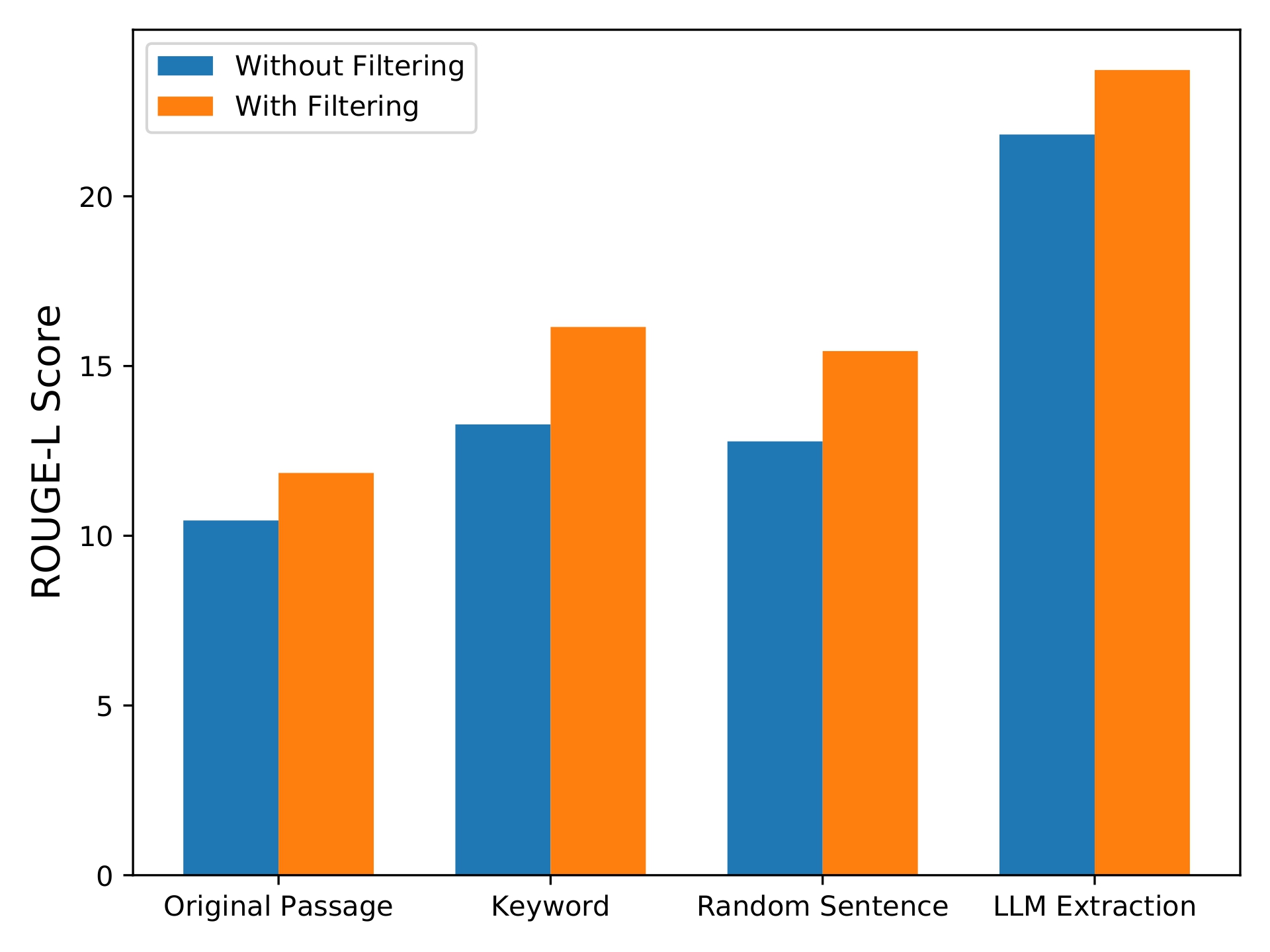}

\end{center}
\caption{\label{fig:extract} 
The performance of different extraction strategies on the SUPERNI benchmark.
We use different extraction strategies to extract fragments from 13,500 unsupervised passages collected from Wikipedia and C4 corpus.
Then we use the instruction generation model trained with seed instructions to generate candidate instructions with these fragments.
}
\end{figure}

%% file: article/experiment.tex
\section{Putting It All Together}
In this section, we combine all the above strategies together to form our final instruction data framework.
Figure~\ref{fig:model} shows the illustration of our framework.
We conduct a comprehensive evaluation to confirm the effectiveness of our framework.
\subsection{Main Results}

To compare with various baselines, we report the performance of our framework in Table~\ref{tab:main}.
We can find that our generated instruction data can outperform all the baselines on SUPERNI and LongForm.
Specifically, only the instruction data generated from an unsupervised corpus can achieve the best performance on LongForm, while the instruction data generated from a supervised corpus can achieve the best performance on SUPERNI.
Combining these two types of data can achieve a balanced result on LongForm and SUPERNI, which can outperform the same amount of the mixing of original in-domain data significantly.

\subsection{GPT-4 Evaluation}
\input{figure/figure_gpt4eval}

To further confirm the effectiveness of our generated data, we follow \citet{vicuna2023} to leverage GPT-4~\citep{openai2023gpt4} to conduct an automatic evaluation.
Specifically, GPT-4 automatically scores the response
quality of different models from 1 to 10 on unseen Vicuna-Instructions. 
Because we do not generate code and math instruction data, we remove evaluation instructions related to code and math and finally keep 70 instructions to test.
The evaluation results are reported in Figure~\ref{fig:eval}, which use that our generated data can outperform Alpaca slightly.
All the evaluation results show the potential of generating high-quality instruction data without using closed-source models.

\subsection{Instruction Diversity Analysis}

\input{figure/fig_diversity_filter}

We also study the diversity of generated instruction data.
Following \citet{wang2022self}, we use the Berkeley Neural Parser~\citep{kitaev2018constituency,kitaev2019multilingual} to
extract the top 20 most common root verbs and their top 4
direct noun objects of our generated instruction data, which are shown in Figure~\ref{fig:diverse}.
From the results, we can observe that the extract-then-generate strategy can improve the diversity of the generated instruction data of the original passages in the unsupervised corpus.

\input{table/table_flant5}

\subsection{Other Backbone Models}
To confirm further the effectiveness of our generated data, we also conduct experiments on T5~\citep{raffel2020exploring} and Flan-T5~\citep{chung2022scaling}, which is reported in Table~\ref{tab:flan}. Without instruction tuning on FLAN collections, T5-XL fails on SUPERNI.
Flan-T5-XL can achieve strong performance on SUPERNI while having a similar result to T5-XL on LongForm.
We speculate that this phenomenon results from there being a large number of classification tasks collected in the FLAN collections, and the outputs of classification tasks are usually too short, which may do little to help improve the long text generation ability.
Despite the weak ability of T5-XL and Flan-T5-XL, we can find that our generated instruction data can achieve comparable and even better performance than the instruction datasets using closed-source models.

%% file: figure/figure_gpt4eval.tex
\begin{figure}[t]
\begin{center}
\includegraphics[width=0.5\textwidth]{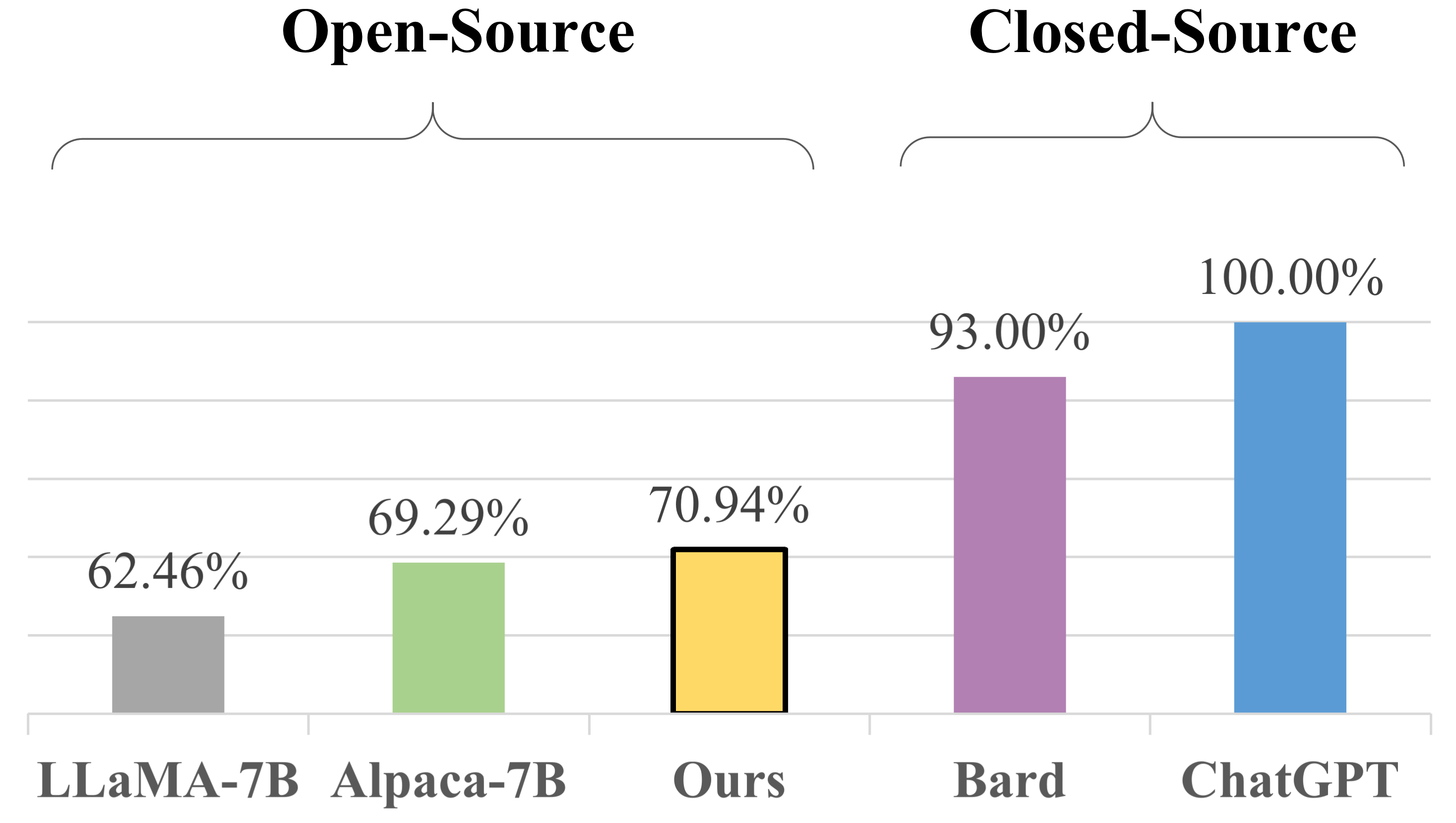}

\end{center}
\caption{\label{fig:eval} 
Relative response quality against ChatGPT, which is assessed by GPT-4.
}

\end{figure}

%% file: figure/fig_diversity_filter.tex
\begin{figure*}[htbp]
	\centering

		\begin{minipage}{0.32\linewidth}
		\centering
		\includegraphics[width=0.9\linewidth]{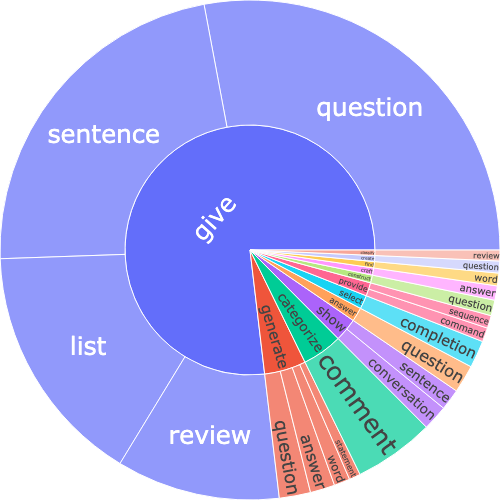}
		\caption*{(a)~SUPERNI}
	\end{minipage}
	\begin{minipage}{0.32\linewidth}
		\centering
		\includegraphics[width=0.9\linewidth]{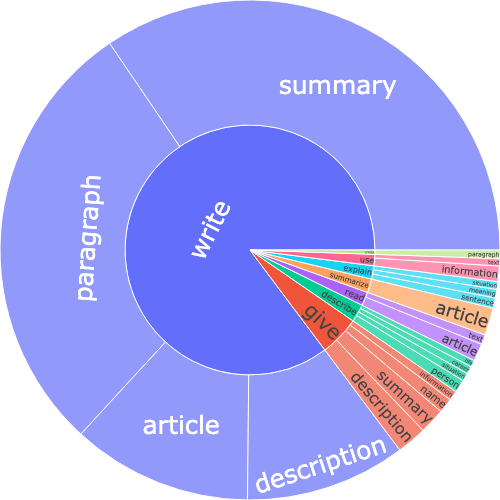}
		\caption*{(b)~Wikipedia\&C4}
	\end{minipage}
	\begin{minipage}{0.32\linewidth}
		\centering
		\includegraphics[width=0.9\linewidth]{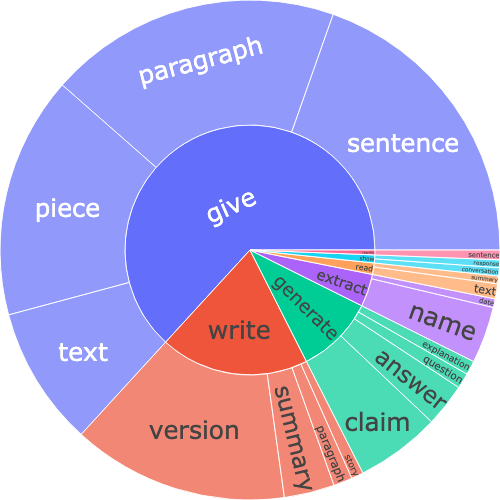}
		\caption*{(c)~Extracted Fragments}
	\end{minipage}
\caption{\label{fig:diverse}The top 20 most common root verbs~(inner circle) and
their top 4 direct noun objects~(outer circle) in the instruction generated from:~\textbf{(a)~SUPERNI};~\textbf{(b)~Wikipedia\&C4};~\textbf{(c)~Fragments Extracted from Wikipedia\&C4}.}
\end{figure*}

%% file: table/table_flant5.tex
\begin{table}[!tbp]
\centering
\resizebox{0.5\textwidth}{!}{
\begin{tabular}{l c c c c c }
\\
%\toprule[1.25pt]
\hline
\bf Method	& \bf SUPERNI & \bf LongForm\\
 \hline
 T5-XL~\citep{raffel2020exploring}&0.89&4.81\\
\quad +Alpaca~\citep{alpaca}&\underline{1.77}&2.50\\
\quad +Dolly~\citep{dolly}&0.96&6.62\\
\quad +DYNASAUR~\citep{yin2023dynosaur}&1.30&2.09\\
\quad +Dromedary~\citep{sun2023principle}&0.84&4.17\\
\quad +SUPERNI~\citep{wang2022super}&1.22&0.79\\
\quad +LongForm~\citep{koksal2023longform}&0.94&\textbf{10.22}\\
\quad +Ours~(Wikipedia\&C4 Source)&1.14&\underline{8.42}\\
\quad +Ours~(SUPERNI Source)&\textbf{1.93}&3.55\\
\quad +Ours~(Wikipedia\&C4+SUPERNI Source)&0.98&4.62\\
\hline
Flan-T5-XL~\citep{chung2022scaling}&36.54&5.11\\
\quad +Alpaca~\citep{alpaca}&22.21&10.38\\
\quad +Dolly~\citep{dolly}&31.21&6.34\\
\quad +DYNASAUR~\citep{yin2023dynosaur}&35.25&5.43\\
\quad +Dromedary~\citep{sun2023principle}&7.83&\textbf{11.64}\\
\quad +SUPERNI~\citep{wang2022super}&29.15&3.63\\
\quad +LongForm~\citep{koksal2023longform}&\textbf{40.37}&7.05\\
\quad +Ours~(Wikipedia\&C4 Source)&34.66&\underline{7.50}\\
\quad +Ours~(SUPERNI Source)&\underline{37.70}&4.83\\
\quad +Ours~(Wikipedia\&C4+SUPERNI Source)&33.78&5.74\\
\hline
\end{tabular}
% \begin{tabular}{l c }
% \\
% %\toprule[1.25pt]
% \toprule
% \bf Method	& \bf SUPERNI \\
%  \midrule
%  T5-XL~\citep{raffel2020exploring}&0.89\\
% \quad +Alpaca~\citep{alpaca}&1.77\\
% \quad +Dolly~\citep{dolly}&0.96\\
% \quad +DYNASAUR~\citep{yin2023dynosaur}&1.30\\
% \quad +Dromedary~\citep{sun2023principle}&0.84\\
% \quad +SUPERNI~\citep{wang2022super}&1.22\\
% \quad +LongForm~\citep{koksal2023longform}&0.94\\
% \quad +Ours~(Wikipedia\&C4 Source)&1.14\\
% \quad +Ours~(SUPERNI Source)&0.93\\
% \quad +Ours~(Wikipedia\&C4+SUPERNI Source)&0.98\\
% \hline
% Flan-T5-XL~\citep{chung2022scaling}&36.54\\
% \quad +Alpaca~\citep{alpaca}&22.21\\
% \quad +Dolly~\citep{dolly}&31.21\\
% \quad +DYNASAUR~\citep{yin2023dynosaur}&35.25\\
% \quad +Dromedary~\citep{sun2023principle}&7.83\\
% \quad +SUPERNI~\citep{wang2022super}&29.15\\
% \quad +LongForm~\citep{koksal2023longform}&\textbf{40.37}\\
% \quad +Ours~(Wikipedia\&C4 Source)&34.66\\
% \quad +Ours~(SUPERNI Source)&\underline{37.70}\\
% \quad +Ours~(Wikipedia\&C4+SUPERNI Source)&33.78\\
% \bottomrule
% \end{tabular}
}
\caption{\label{tab:flan} 
The performance of our generated instruction data and all instruction dataset baselines on SUPERNI benchmarks based on T5-XL and Flan-T5-XL.
\textbf{\textit{Bold}} and \textit{\underline{Underline}} indicate the best and the second best performance respectively.
Following the original usage of the automatic metric of benchmarks~\citep{wang2022super,koksal2023longform}, we report the ROUGE-L~\citep{lin2004rouge} score for SUPERNI. 
}
% \vspace{-0.4cm}
\end{table}

%% file: article/relatedwork.tex
\section{Related Work}

In recent years, instruction tuning has gained significant attention, which can enable LLMs to generalize to unseen tasks.
In this section, we divide existing instruction datasets into two groups: manual annotation instruction datasets and model generation instruction datasets.
In this section, we introduce these two types of methods.

\paragraph{Manual Annotation Instruction Datasets}
For the purpose of generalizing LLMs to unseen tasks, instruct tuning needs a large number of instruction data with various task types.
The most straightforward way to achieve this goal is to collect from existing NLP datasets.
Therefore, early instruction datasets are collected from large-scale existing NLP task datasets and transformed into instruction formats with manual written templates, e.g., Natural Instructions~\citep{mishra2021cross}, Flan 2021~\citep{wei2021finetuned}, and T0~\citep{sanh2021multitask}. 
With the scaling law of instruction data that more task types and data can improve the performance of instruction tuning~\citep{xu2022zeroprompt}, subsequent works collect more data from existing datasets to expand tasks to the thousands, e.g., ZeroPrompt~\citep{xu2022zeroprompt}, Super-Natural Instructions~\citep{wang2022super} and OPT-IML~\citep{iyer2022opt}.
To further enrich data sources, xP3~\citep{muennighoff2022crosslingual} adds multilingual instruction tuning;
Flan Collection~\citep{chung2022scaling,longpre2023flan} collects chain-of-thought data and samples some data to be organized in the format of in-context learning.
\citet{gu2022learning} leverages human-written rules and templates to convert unsupervised data into the instruction data format to further raise the number of data resources.
In order to align with human requirements in realistic scenarios rather than collecting instruction data from existing NLP datasets, some recent works focus on collecting instruction data from realistic scenarios~\citep{ouyang2022training,dolly}.

\paragraph{Model Generation Instruction Datasets}
Limited by the high cost of manual annotation, it is unaffordable to collect high-quality instruction data manually.
Therefore, model generation models focus on generating instruction data automatically.
Existing methods have explored the effectiveness of employing large language models to generate high-quality instruction data. 
Specifically, by providing multiple seed instruction-tuning tasks as prompts, Self-Instruct~\citep{wang2022self}, Unnatural Instruction~\citep{honovich2022unnatural}, and Alpaca~\citep{alpaca} can follow the format of the given seed tasks to generate instruction data automatically.
Subsequently, there is a line of work to improve the quality of generated data further.
Specifically, \citet{xu2023wizardlm} focuses on improving the task complexity of instruction data;
\citet{jiang2023lion} proposes an adversarial distillation framework to utilize the feedback of student models;
\citet{ding2023enhancing} utilizes two separate LLM APIs to generate conversational instruction data.
Besides, relying solely on LLMs to generate both instruction and outputs may generate low-quality data~\citep{koksal2023longform,yang2023refgpt}.
To improve the quality of generated instruction data, LongForm~\citep{koksal2023longform} is the pioneering work leveraging LLMs to generate the instruction based on an existing corpus.
Dynasaur~\citep{yin2023dynosaur} uses LLMs to generate instruction data with the help of the meta-information of existing NLP datasets.
RefGPT~\citep{yang2023refgpt} handles the hallucination problem by constraining the LLMs to rely on the provided reference instead of generating dialogues based solely on their own knowledge.

% \subsection{Task-Specific Instruction Tuning Methods}

% \subsection{General-Purpose Instruction Tuning Methods}

% \subsection{Data Generation Methods}

%% file: article/conclusion.tex
\section{Conclusion and Future Work}
Instruction tuning has gained more and more attention, which uses high-quality instruction data to enable LLMs to generalize to various tasks.
Limited by the high cost, manual annotation instruction datasets lack a large amount of high-quality instruction data.
Therefore, recent works focus on leveraging powerful closed-source models to generate instruction data automatically, which have achieved huge success.
However, the usage requirements of closed-source models 
are usually forbidden to use their outputs to develop machine learning models, which causes these methods potential risks.

In this work, we first verify the potential of generating high-quality instruction datasets without using closed-source models by studying different strategies.
Subsequently, we introduce a novel instruction data generation framework, consisting of the best variants of existing methods and our novel proposed strategies.
Experimental results confirm the effectiveness of our data, which can outperform Alpaca in both benchmark and GPT-4 evaluation.

Despite the promising results achieved in this work, it is still a large performance gap between closed-source models and open-source models.
Therefore, we hope this work can prompt the development of open-source instruction tuning models to narrow the gap with closed-source models continuously.
Besides, this work is a preliminary study to explore the potential of generating high-quality instruction datasets without using closed-source models.
In future work, we will try to generate instructions with more diverse and complex task types, such as math, code, etc.
We will also generalize our framework to support more languages.

%% file: figure/fig_benchmark.tex
\begin{figure*}[h]
\begin{minipage}[t]{0.4\textwidth}
\centering
\includegraphics[width=0.95\textwidth]{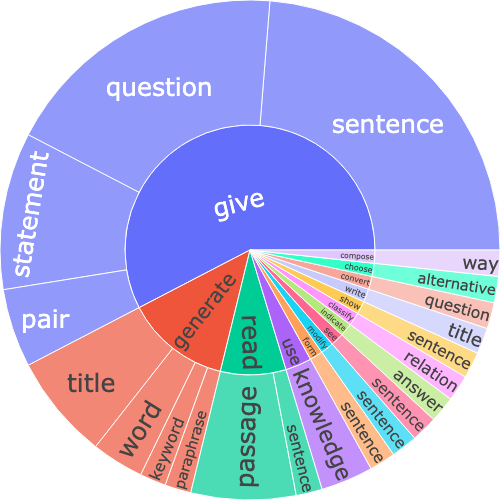}
\caption*{(a)~SUPERNI test set.}
\label{fig:side:a}
% \vspace{0.5cm}
\end{minipage}%
\hspace{0.5cm}
\begin{minipage}[t]{0.4\textwidth}
\centering
\includegraphics[width=0.95\textwidth]{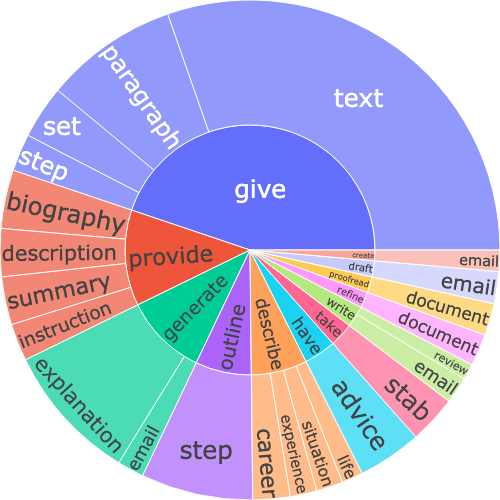}
\caption*{(b)~LongForm test set.}
\label{fig:side:b}
% \vspace{0.5cm}
\end{minipage}
\centering
\caption{\label{fig:diverse_test}The top 20 most common root verbs~(inner circle) and
their top 4 direct noun objects~(outer circle) in the instruction generated from:\textbf{(a)~SUPERNI test set};\textbf{(b)~LongForm test set}.}
\end{figure*}

%% file: table/prompt.tex
\begin{table*}[ht]
\centering
\small
\renewcommand\arraystretch{1.6}
% \resizebox{\columnwidth}{!}{

\begin{tabular}{p{14cm}}
\\
%\toprule[1.25pt]
\toprule
\makecell[c]{\bf Instruction with Input} \\
\midrule

        Below is an instruction that describes a task, paired with an input that provides further context.Write a response that appropriately completes the request.
\\
  \\      
        \#\#\# Instruction:\\
        \textcolor{red}{\{\{ instruction \}\}}\\
    \\    
        \#\#\# Input:\\
       \textcolor{red}{\{\{input\}\}}\\  
       \\
        \#\#\# Response:\\

\midrule
 \makecell[c]{\bf Instruction without Input} \\
\midrule
Below is an instruction that describes a task. Write a response that appropriately completes the request.
\\
  \\      
        \#\#\# Instruction:\\
        \textcolor{red}{\{\{ instruction \}\}}\\
       \\
        \#\#\# Response:\\

\bottomrule
\end{tabular}%
\caption{\label{tab:alpaca_insturction} The instruction following prompt of Alpaca.}

% }
% \vspace{-0.4cm}
\end{table*}

%% file: table/prompt1.tex
\begin{table*}[ht]
\centering
\small
\renewcommand\arraystretch{1.6}
% \resizebox{\columnwidth}{!}{

\begin{tabular}{p{14cm}}
\\
%\toprule[1.25pt]
\toprule
\makecell[c]{\bf Instruction Date Generation Meta Prompt} \\
\midrule

You are asked to come up with a set of 20 diverse task instructions. These task instructions will be given to a GPT model and we will evaluate the GPT model for completing the instructions.

Here are the requirements:  

1. Try not to repeat the verb for each instruction to maximize diversity.  

2. The language used for the instruction also should be diverse. For example, you should combine questions with imperative instrucitons.

3. The type of instructions should be diverse. The list should include diverse types of tasks like open-ended generation, classification, editing, etc.

4. A GPT language model should be able to complete the instruction. For example, do not ask the assistant to create any visual or audio output. For another example, do not ask the assistant to wake you up at 5pm or set a reminder because it cannot perform any action.

5. The instructions should be in English.

6. The instructions should be 1 to 2 sentences long. Either an imperative sentence or a question is permitted.

7. You should generate an appropriate input to the instruction. The input field should contain a specific example provided for the instruction. It should involve realistic data and should not contain simple placeholders. The input should provide substantial content to make the instruction challenging but should ideally not exceed 100 words.

8. Not all instructions require input. For example, when a instruction asks about some general information, "what is the highest peak in the world", it is not necssary to provide a specific context. In this case, we simply put "<noinput>" in the input field.

9. The output should be an appropriate response to the instruction and the input. Make sure the output is less than 100 words.

List of 20 tasks:

1.\, Instruction:

\{instruction\_example1\}

1.\, Input:

\{input\_example1\}

1.\, Output:

\{output\_example1\}

2.\, Instruction:

\{instruction\_example2\}

2.\, Input:

\{input\_example2\}

2.\, Output:

\{output\_example2\}

3.\, Instruction:

\{instruction\_example3\}

3.\, Input:

\{input\_example3\}

3.\, Output:

\{output\_example3\}

4.\, Instruction:\\

\bottomrule

\end{tabular}%
\caption{\label{tab:alpacainstruction} The instruction data generation prompt of Alpaca.}

% }
% \vspace{-0.4cm}
\end{table*}

%% file: table/prompt2.tex
\begin{table*}[ht]
\centering
\small
\renewcommand\arraystretch{1.6}
% \resizebox{\columnwidth}{!}{

\begin{tabular}{p{14cm}}
\\
%\toprule[1.25pt]
\toprule
\makecell[c]{\bf Template for the Instruction Style} \\
\midrule

Instruction: X

Output: <corpus\_example>

What kind of instruction could this be the answer to?

X:\\

\midrule
 \makecell[c]{\bf Template for the Informal Chatbot Style } \\
\midrule
You are a chatbot. A user sent you an informal message
and your reply is as follows.

Message: X

Reply: <corpus\_example>

What is the informal message X?

X:
 \\
\midrule
 \makecell[c]{\bf  Template for the Search Engine/query Style} \\
\midrule
You are a search engine. A person queried something in
detail and the most relevant document about the query is
as follows.

Query: X

Document: <corpus\_example>

What is the detailed query X?

X:
 \\
\bottomrule
\end{tabular}%
\caption{\label{tab:longform_generation} The instruction data generation prompt of LongForm. 
<corpus\_example> represents the selected passages.}

% }
% \vspace{-0.4cm}
\end{table*}

%% file: table/table_extractprompt.tex
\begin{table*}[ht]
\centering
\small
\renewcommand\arraystretch{1.6}
% \resizebox{\columnwidth}{!}{

\begin{tabular}{p{14cm}}
\\
%\toprule[1.25pt]
\toprule

        Below is an instruction that describes a task, paired with an input that provides further context.
        Write a response that appropriately completes the request.
\\
  \\      
        \#\#\# Instruction:\\
        Select key segments from the given article below.\\
    \\    
        \#\#\# Input:\\
       \textcolor{red}{\{\{input\}\}}\\  
       \\
        \#\#\# Response:\\
\bottomrule
\end{tabular}%
\caption{\label{tab:llmextract} The LLM extraction prompt.}

% }
% \vspace{-0.4cm}
\end{table*}

%% file: main.bbl
\begin{thebibliography}{45}
\expandafter\ifx\csname natexlab\endcsname\relax\def\natexlab#1{#1}\fi

\bibitem[{Banerjee and Lavie(2005)}]{banerjee2005meteor}
Satanjeev Banerjee and Alon Lavie. 2005.
\newblock Meteor: An automatic metric for mt evaluation with improved
  correlation with human judgments.
\newblock In \emph{Proceedings of the acl workshop on intrinsic and extrinsic
  evaluation measures for machine translation and/or summarization}, pages
  65--72.

\bibitem[{Brown et~al.(2020)Brown, Mann, Ryder, Subbiah, Kaplan, Dhariwal,
  Neelakantan, Shyam, Sastry, Askell et~al.}]{brown2020language}
Tom Brown, Benjamin Mann, Nick Ryder, Melanie Subbiah, Jared~D Kaplan, Prafulla
  Dhariwal, Arvind Neelakantan, Pranav Shyam, Girish Sastry, Amanda Askell,
  et~al. 2020.
\newblock Language models are few-shot learners.
\newblock \emph{Advances in neural information processing systems},
  33:1877--1901.

\bibitem[{Bryant et~al.(2019)Bryant, Felice, Andersen, and
  Briscoe}]{bryant2019bea}
Christopher Bryant, Mariano Felice, {\O}istein~E Andersen, and Ted Briscoe.
  2019.
\newblock The bea-2019 shared task on grammatical error correction.
\newblock In \emph{Proceedings of the Fourteenth Workshop on Innovative Use of
  NLP for Building Educational Applications}, pages 52--75.

\bibitem[{Campos et~al.(2020)Campos, Mangaravite, Pasquali, Jorge, Nunes, and
  Jatowt}]{campos2020yake}
Ricardo Campos, V{\'\i}tor Mangaravite, Arian Pasquali, Al{\'\i}pio Jorge,
  C{\'e}lia Nunes, and Adam Jatowt. 2020.
\newblock Yake! keyword extraction from single documents using multiple local
  features.
\newblock \emph{Information Sciences}, 509:257--289.

\bibitem[{Campos et~al.(2018{\natexlab{a}})Campos, Mangaravite, Pasquali,
  Jorge, Nunes, and Jatowt}]{campos2018text}
Ricardo Campos, V{\'\i}tor Mangaravite, Arian Pasquali, Al{\'\i}pio~M{\'a}rio
  Jorge, C{\'e}lia Nunes, and Adam Jatowt. 2018{\natexlab{a}}.
\newblock A text feature based automatic keyword extraction method for single
  documents.
\newblock In \emph{European conference on information retrieval}, pages
  684--691. Springer.

\bibitem[{Campos et~al.(2018{\natexlab{b}})Campos, Mangaravite, Pasquali,
  Jorge, Nunes, and Jatowt}]{campos2018yake}
Ricardo Campos, V{\'\i}tor Mangaravite, Arian Pasquali, Al{\'\i}pio~M{\'a}rio
  Jorge, C{\'e}lia Nunes, and Adam Jatowt. 2018{\natexlab{b}}.
\newblock Yake! collection-independent automatic keyword extractor.
\newblock In \emph{European Conference on Information Retrieval}, pages
  806--810. Springer.

\bibitem[{Chen et~al.(2023)Chen, Zhang, Zhang, Yang, Hu, Ma, Yanggong, and
  Zhao}]{chen2023maybe}
Hao Chen, Yiming Zhang, Qi~Zhang, Hantao Yang, Xiaomeng Hu, Xuetao Ma, Yifan
  Yanggong, and Junbo Zhao. 2023.
\newblock Maybe only 0.5\% data is needed: A preliminary exploration of low
  training data instruction tuning.
\newblock \emph{arXiv preprint arXiv:2305.09246}.

\bibitem[{Chiang et~al.(2023)Chiang, Li, Lin, Sheng, Wu, Zhang, Zheng, Zhuang,
  Zhuang, Gonzalez, Stoica, and Xing}]{vicuna2023}
Wei-Lin Chiang, Zhuohan Li, Zi~Lin, Ying Sheng, Zhanghao Wu, Hao Zhang, Lianmin
  Zheng, Siyuan Zhuang, Yonghao Zhuang, Joseph~E. Gonzalez, Ion Stoica, and
  Eric~P. Xing. 2023.
\newblock \href {https://lmsys.org/blog/2023-03-30-vicuna/} {Vicuna: An
  open-source chatbot impressing gpt-4 with 90\%* chatgpt quality}.

\bibitem[{Chung et~al.(2022)Chung, Hou, Longpre, Zoph, Tay, Fedus, Li, Wang,
  Dehghani, Brahma et~al.}]{chung2022scaling}
Hyung~Won Chung, Le~Hou, Shayne Longpre, Barret Zoph, Yi~Tay, William Fedus,
  Eric Li, Xuezhi Wang, Mostafa Dehghani, Siddhartha Brahma, et~al. 2022.
\newblock Scaling instruction-finetuned language models.
\newblock \emph{arXiv preprint arXiv:2210.11416}.

\bibitem[{Cohen(2015)}]{enron}
William~W. Cohen. 2015.
\newblock Enron email dataset.
\newblock \url{https://www.cs.cmu.edu/~enron/}.

\bibitem[{Databricks(2023)}]{dolly}
Databricks. 2023.
\newblock Databricks’ dolly, a large language model trained on the databricks
  machine learning platform.
\newblock \url{https://github.com/databrickslabs/dolly}.

\bibitem[{Ding et~al.(2023)Ding, Chen, Xu, Qin, Zheng, Hu, Liu, Sun, and
  Zhou}]{ding2023enhancing}
Ning Ding, Yulin Chen, Bokai Xu, Yujia Qin, Zhi Zheng, Shengding Hu, Zhiyuan
  Liu, Maosong Sun, and Bowen Zhou. 2023.
\newblock Enhancing chat language models by scaling high-quality instructional
  conversations.
\newblock \emph{arXiv preprint arXiv:2305.14233}.

\bibitem[{Gao et~al.(2020)Gao, Biderman, Black, Golding, Hoppe, Foster, Phang,
  He, Thite, Nabeshima et~al.}]{gao2020pile}
Leo Gao, Stella Biderman, Sid Black, Laurence Golding, Travis Hoppe, Charles
  Foster, Jason Phang, Horace He, Anish Thite, Noa Nabeshima, et~al. 2020.
\newblock The pile: An 800gb dataset of diverse text for language modeling.
\newblock \emph{arXiv preprint arXiv:2101.00027}.

\bibitem[{Gu et~al.(2022)Gu, Ke, Zhu, and Huang}]{gu2022learning}
Yuxian Gu, Pei Ke, Xiaoyan Zhu, and Minlie Huang. 2022.
\newblock Learning instructions with unlabeled data for zero-shot cross-task
  generalization.
\newblock \emph{arXiv preprint arXiv:2210.09175}.

\bibitem[{Holtzman et~al.(2019)Holtzman, Buys, Du, Forbes, and
  Choi}]{holtzman2019curious}
Ari Holtzman, Jan Buys, Li~Du, Maxwell Forbes, and Yejin Choi. 2019.
\newblock The curious case of neural text degeneration.
\newblock In \emph{International Conference on Learning Representations}.

\bibitem[{Honovich et~al.(2022)Honovich, Scialom, Levy, and
  Schick}]{honovich2022unnatural}
Or~Honovich, Thomas Scialom, Omer Levy, and Timo Schick. 2022.
\newblock Unnatural instructions: Tuning language models with (almost) no human
  labor.
\newblock \emph{arXiv preprint arXiv:2212.09689}.

\bibitem[{Iyer et~al.(2022)Iyer, Lin, Pasunuru, Mihaylov, Simig, Yu, Shuster,
  Wang, Liu, Koura et~al.}]{iyer2022opt}
Srinivasan Iyer, Xi~Victoria Lin, Ramakanth Pasunuru, Todor Mihaylov,
  D{\'a}niel Simig, Ping Yu, Kurt Shuster, Tianlu Wang, Qing Liu, Punit~Singh
  Koura, et~al. 2022.
\newblock Opt-iml: Scaling language model instruction meta learning through the
  lens of generalization.
\newblock \emph{arXiv preprint arXiv:2212.12017}.

\bibitem[{Jiang et~al.(2023)Jiang, Chan, Chen, and Wang}]{jiang2023lion}
Yuxin Jiang, Chunkit Chan, Mingyang Chen, and Wei Wang. 2023.
\newblock Lion: Adversarial distillation of closed-source large language model.
\newblock \emph{arXiv preprint arXiv:2305.12870}.

\bibitem[{Kitaev et~al.(2019)Kitaev, Cao, and Klein}]{kitaev2019multilingual}
Nikita Kitaev, Steven Cao, and Dan Klein. 2019.
\newblock Multilingual constituency parsing with self-attention and
  pre-training.
\newblock In \emph{Proceedings of the 57th Annual Meeting of the Association
  for Computational Linguistics}, pages 3499--3505.

\bibitem[{Kitaev and Klein(2018)}]{kitaev2018constituency}
Nikita Kitaev and Dan Klein. 2018.
\newblock Constituency parsing with a self-attentive encoder.
\newblock In \emph{Proceedings of the 56th Annual Meeting of the Association
  for Computational Linguistics (Volume 1: Long Papers)}, pages 2676--2686.

\bibitem[{K{\"o}ksal et~al.(2023)K{\"o}ksal, Schick, Korhonen, and
  Sch{\"u}tze}]{koksal2023longform}
Abdullatif K{\"o}ksal, Timo Schick, Anna Korhonen, and Hinrich Sch{\"u}tze.
  2023.
\newblock Longform: Optimizing instruction tuning for long text generation with
  corpus extraction.
\newblock \emph{arXiv preprint arXiv:2304.08460}.

\bibitem[{Koupaee and Wang(2018)}]{koupaee2018wikihow}
Mahnaz Koupaee and William~Yang Wang. 2018.
\newblock Wikihow: A large scale text summarization dataset.
\newblock \emph{arXiv preprint arXiv:1810.09305}.

\bibitem[{Lin(2004)}]{lin2004rouge}
Chin-Yew Lin. 2004.
\newblock Rouge: A package for automatic evaluation of summaries.
\newblock In \emph{Text summarization branches out}, pages 74--81.

\bibitem[{Longpre et~al.(2023)Longpre, Hou, Vu, Webson, Chung, Tay, Zhou, Le,
  Zoph, Wei et~al.}]{longpre2023flan}
Shayne Longpre, Le~Hou, Tu~Vu, Albert Webson, Hyung~Won Chung, Yi~Tay, Denny
  Zhou, Quoc~V Le, Barret Zoph, Jason Wei, et~al. 2023.
\newblock The flan collection: Designing data and methods for effective
  instruction tuning.
\newblock \emph{arXiv preprint arXiv:2301.13688}.

\bibitem[{Loshchilov and Hutter(2019)}]{loshchilov2018decoupled}
Ilya Loshchilov and Frank Hutter. 2019.
\newblock \href {https://openreview.net/forum?id=Bkg6RiCqY7} {Decoupled weight
  decay regularization}.
\newblock In \emph{International Conference on Learning Representations}.

\bibitem[{Mishra et~al.(2022)Mishra, Khashabi, Baral, and
  Hajishirzi}]{mishra2021cross}
Swaroop Mishra, Daniel Khashabi, Chitta Baral, and Hannaneh Hajishirzi. 2022.
\newblock \href {https://doi.org/10.18653/v1/2022.acl-long.244} {Cross-task
  generalization via natural language crowdsourcing instructions}.
\newblock In \emph{Proceedings of the 60th Annual Meeting of the Association
  for Computational Linguistics (Volume 1: Long Papers)}, pages 3470--3487,
  Dublin, Ireland. Association for Computational Linguistics.

\bibitem[{Muennighoff et~al.(2022)Muennighoff, Wang, Sutawika, Roberts,
  Biderman, Scao, Bari, Shen, Yong, Schoelkopf
  et~al.}]{muennighoff2022crosslingual}
Niklas Muennighoff, Thomas Wang, Lintang Sutawika, Adam Roberts, Stella
  Biderman, Teven~Le Scao, M~Saiful Bari, Sheng Shen, Zheng-Xin Yong, Hailey
  Schoelkopf, et~al. 2022.
\newblock Crosslingual generalization through multitask finetuning.
\newblock \emph{arXiv preprint arXiv:2211.01786}.

\bibitem[{OpenAI(2023)}]{openai2023gpt4}
OpenAI. 2023.
\newblock \href {http://arxiv.org/abs/2303.08774} {Gpt-4 technical report}.

\bibitem[{Ouyang et~al.(2022)Ouyang, Wu, Jiang, Almeida, Wainwright, Mishkin,
  Zhang, Agarwal, Slama, Ray et~al.}]{ouyang2022training}
Long Ouyang, Jeffrey Wu, Xu~Jiang, Diogo Almeida, Carroll Wainwright, Pamela
  Mishkin, Chong Zhang, Sandhini Agarwal, Katarina Slama, Alex Ray, et~al.
  2022.
\newblock Training language models to follow instructions with human feedback.
\newblock \emph{Advances in Neural Information Processing Systems},
  35:27730--27744.

\bibitem[{Peng et~al.(2023)Peng, Li, He, Galley, and Gao}]{peng2023instruction}
Baolin Peng, Chunyuan Li, Pengcheng He, Michel Galley, and Jianfeng Gao. 2023.
\newblock Instruction tuning with gpt-4.
\newblock \emph{arXiv preprint arXiv:2304.03277}.

\bibitem[{Raffel et~al.(2020)Raffel, Shazeer, Roberts, Lee, Narang, Matena,
  Zhou, Li, and Liu}]{raffel2020exploring}
Colin Raffel, Noam Shazeer, Adam Roberts, Katherine Lee, Sharan Narang, Michael
  Matena, Yanqi Zhou, Wei Li, and Peter~J Liu. 2020.
\newblock Exploring the limits of transfer learning with a unified text-to-text
  transformer.
\newblock \emph{The Journal of Machine Learning Research}, 21(1):5485--5551.

\bibitem[{Sanh et~al.(2021)Sanh, Webson, Raffel, Bach, Sutawika, Alyafeai,
  Chaffin, Stiegler, Scao, Raja et~al.}]{sanh2021multitask}
Victor Sanh, Albert Webson, Colin Raffel, Stephen~H Bach, Lintang Sutawika,
  Zaid Alyafeai, Antoine Chaffin, Arnaud Stiegler, Teven~Le Scao, Arun Raja,
  et~al. 2021.
\newblock Multitask prompted training enables zero-shot task generalization.
\newblock \emph{arXiv preprint arXiv:2110.08207}.

\bibitem[{Srivastava et~al.(2022)Srivastava, Rastogi, Rao, Shoeb, Abid, Fisch,
  Brown, Santoro, Gupta, Garriga-Alonso et~al.}]{srivastava2022beyond}
Aarohi Srivastava, Abhinav Rastogi, Abhishek Rao, Abu Awal~Md Shoeb, Abubakar
  Abid, Adam Fisch, Adam~R Brown, Adam Santoro, Aditya Gupta, Adri{\`a}
  Garriga-Alonso, et~al. 2022.
\newblock Beyond the imitation game: Quantifying and extrapolating the
  capabilities of language models.
\newblock \emph{arXiv preprint arXiv:2206.04615}.

\bibitem[{Sun et~al.(2023)Sun, Shen, Zhou, Zhang, Chen, Cox, Yang, and
  Gan}]{sun2023principle}
Zhiqing Sun, Yikang Shen, Qinhong Zhou, Hongxin Zhang, Zhenfang Chen, David
  Cox, Yiming Yang, and Chuang Gan. 2023.
\newblock Principle-driven self-alignment of language models from scratch with
  minimal human supervision.
\newblock \emph{arXiv preprint arXiv:2305.03047}.

\bibitem[{Taori et~al.(2023)Taori, Gulrajani, Zhang, Dubois, Li, Guestrin,
  Liang, and Hashimoto}]{alpaca}
Rohan Taori, Ishaan Gulrajani, Tianyi Zhang, Yann Dubois, Xuechen Li, Carlos
  Guestrin, Percy Liang, and Tatsunori~B. Hashimoto. 2023.
\newblock Stanford alpaca: An instruction-following llama model.
\newblock \url{https://github.com/tatsu-lab/stanford_alpaca}.

\bibitem[{Touvron et~al.(2023)Touvron, Lavril, Izacard, Martinet, Lachaux,
  Lacroix, Rozi{\`e}re, Goyal, Hambro, Azhar et~al.}]{touvron2023llama}
Hugo Touvron, Thibaut Lavril, Gautier Izacard, Xavier Martinet, Marie-Anne
  Lachaux, Timoth{\'e}e Lacroix, Baptiste Rozi{\`e}re, Naman Goyal, Eric
  Hambro, Faisal Azhar, et~al. 2023.
\newblock Llama: Open and efficient foundation language models.
\newblock \emph{arXiv preprint arXiv:2302.13971}.

\bibitem[{Wang et~al.(2022{\natexlab{a}})Wang, Kordi, Mishra, Liu, Smith,
  Khashabi, and Hajishirzi}]{wang2022self}
Yizhong Wang, Yeganeh Kordi, Swaroop Mishra, Alisa Liu, Noah~A Smith, Daniel
  Khashabi, and Hannaneh Hajishirzi. 2022{\natexlab{a}}.
\newblock Self-instruct: Aligning language model with self generated
  instructions.
\newblock \emph{arXiv preprint arXiv:2212.10560}.

\bibitem[{Wang et~al.(2022{\natexlab{b}})Wang, Mishra, Alipoormolabashi, Kordi,
  Mirzaei, Naik, Ashok, Dhanasekaran, Arunkumar, Stap et~al.}]{wang2022super}
Yizhong Wang, Swaroop Mishra, Pegah Alipoormolabashi, Yeganeh Kordi, Amirreza
  Mirzaei, Atharva Naik, Arjun Ashok, Arut~Selvan Dhanasekaran, Anjana
  Arunkumar, David Stap, et~al. 2022{\natexlab{b}}.
\newblock Super-naturalinstructions: Generalization via declarative
  instructions on 1600+ nlp tasks.
\newblock In \emph{Proceedings of the 2022 Conference on Empirical Methods in
  Natural Language Processing}, pages 5085--5109.

\bibitem[{Wei et~al.(2021)Wei, Bosma, Zhao, Guu, Yu, Lester, Du, Dai, and
  Le}]{wei2021finetuned}
Jason Wei, Maarten Bosma, Vincent~Y Zhao, Kelvin Guu, Adams~Wei Yu, Brian
  Lester, Nan Du, Andrew~M Dai, and Quoc~V Le. 2021.
\newblock Finetuned language models are zero-shot learners.
\newblock \emph{arXiv preprint arXiv:2109.01652}.

\bibitem[{Wolf et~al.(2020)Wolf, Debut, Sanh, Chaumond, Delangue, Moi, Cistac,
  Rault, Louf, Funtowicz et~al.}]{wolf2020transformers}
Thomas Wolf, Lysandre Debut, Victor Sanh, Julien Chaumond, Clement Delangue,
  Anthony Moi, Pierric Cistac, Tim Rault, R{\'e}mi Louf, Morgan Funtowicz,
  et~al. 2020.
\newblock Transformers: State-of-the-art natural language processing.
\newblock In \emph{Proceedings of the 2020 conference on empirical methods in
  natural language processing: system demonstrations}, pages 38--45.

\bibitem[{Xu et~al.(2023)Xu, Sun, Zheng, Geng, Zhao, Feng, Tao, and
  Jiang}]{xu2023wizardlm}
Can Xu, Qingfeng Sun, Kai Zheng, Xiubo Geng, Pu~Zhao, Jiazhan Feng, Chongyang
  Tao, and Daxin Jiang. 2023.
\newblock \href {http://arxiv.org/abs/2304.12244} {Wizardlm: Empowering large
  language models to follow complex instructions}.

\bibitem[{Xu et~al.(2022)Xu, Chen, Du, Shao, Wang, Li, and
  Yang}]{xu2022zeroprompt}
Hanwei Xu, Yujun Chen, Yulun Du, Nan Shao, Yanggang Wang, Haiyu Li, and Zhilin
  Yang. 2022.
\newblock Zeroprompt: Scaling prompt-based pretraining to 1,000 tasks improves
  zero-shot generalization.
\newblock \emph{arXiv preprint arXiv:2201.06910}.

\bibitem[{Yang et~al.(2023)Yang, Yuan, Fan, Yang, Wang, Wang, and
  Zhao}]{yang2023refgpt}
Dongjie Yang, Ruifeng Yuan, YuanTao Fan, YiFei Yang, Zili Wang, Shushen Wang,
  and Hai Zhao. 2023.
\newblock Refgpt: Reference-> truthful \& customized dialogues generation by
  gpts and for gpts.
\newblock \emph{arXiv preprint arXiv:2305.14994}.

\bibitem[{Yin et~al.(2023)Yin, Liu, Yin, Zhong, Bansal, Han, and
  Chang}]{yin2023dynosaur}
Da~Yin, Xiao Liu, Fan Yin, Ming Zhong, Hritik Bansal, Jiawei Han, and Kai-Wei
  Chang. 2023.
\newblock Dynosaur: A dynamic growth paradigm for instruction-tuning data
  curation.
\newblock \emph{arXiv preprint arXiv:2305.14327}.

\bibitem[{Zhou et~al.(2023)Zhou, Liu, Xu, Iyer, Sun, Mao, Ma, Efrat, Yu, Yu
  et~al.}]{zhou2023lima}
Chunting Zhou, Pengfei Liu, Puxin Xu, Srini Iyer, Jiao Sun, Yuning Mao, Xuezhe
  Ma, Avia Efrat, Ping Yu, Lili Yu, et~al. 2023.
\newblock Lima: Less is more for alignment.
\newblock \emph{arXiv preprint arXiv:2305.11206}.

\end{thebibliography}
